\documentclass[12pt]{article}
\usepackage{graphicx}
\usepackage{multirow}
\usepackage{booktabs}
\usepackage{amsmath}
\usepackage[a4paper, total={6in, 9in}]{geometry}
\usepackage[font=footnotesize]{caption}
\usepackage{hyperref}
\usepackage{lineno}
\usepackage{makecell}
\usepackage[ruled]{algorithm2e}
\usepackage{algpseudocode}
\begin{document}
\title{Adaptive Reorganization of Neural Pathways for Continual Learning with  Spiking Neural Networks}

\author
{Bing Han$^{1,3,\#}$, Feifei Zhao$^{1,\#}$, Wenxuan Pan$^{1,3}$, \\Zhuoya Zhao$^{1}$, Qingqun Kong$^{1,3}$, Yi Zeng$^{1,2,3,4,*}$\\
\\
\normalsize{$^{1}$Brain-inspired Cognitive Intelligence Lab, }\\
\normalsize{Institute of Automation,Chinese Academy of Sciences}\\
\normalsize{$^{2}$State Key Laboratory of Brain Cognition and Brain-inspired Intelligence Technology, }\\
\normalsize{Chinese Academy of Sciences}\\
\normalsize{$^{3}$ School of Artificial Intelligence, University of Chinese Academy of Sciences}\\
\normalsize{$^{4}$ Center for Long-term Artificial Intelligence}\\
\normalsize{$^{*}$Corresponding authors: yi.zeng@ia.ac.cn}\\
\normalsize{$^{\#}$Co-first authors with equal contribution}\\
}

\maketitle


\begin{abstract}
The human brain can adaptive reorganization rich and diverse sparse neural pathways to incrementally master hundreds of cognitive tasks. However, most existing continual learning algorithms for deep artificial and spiking neural networks are unable to adequately auto-regulate the limited resources in the network, which leads to performance drop along with energy consumption rise as the increase of tasks. In this paper, we propose a brain-inspired continual learning algorithm with adaptive reorganization of neural pathways, which employs Adaptive Dynamic Routing Reorganization Network (ADR-SNN) to reorganize the single and limited Spiking Neural Network into rich sparse neural pathways to efficiently cope with incremental tasks. The proposed model demonstrates consistent superiority in performance, energy consumption, and memory capacity on diverse continual learning tasks ranging from child-like simple to complex tasks, as well as on generalized CIFADR100 and ImageNet datasets. In particular, the ADR-SNN model excels at learning more complex tasks as well as more tasks, and is able to integrate the past learned knowledge with the information from the current task, showing the backward transfer ability to facilitate the old tasks. Meanwhile, the proposed model exhibits self-repairing ability to irreversible damage and for pruned networks, could automatically allocate new pathway from the retained network to recover memory for forgotten knowledge.
  
\end{abstract}

\section*{Keywords}
Adaptive Reorganization Regulation, Continual Learning,  Reorganize Sparse Neural Pathways,  Spiking Neural Networks,  Child-like Simple-to-complex Cognitive Tasks  

\section{Introduction}
\begin{figure}[htbp]
	\centering 
	\includegraphics[width=0.6\linewidth]{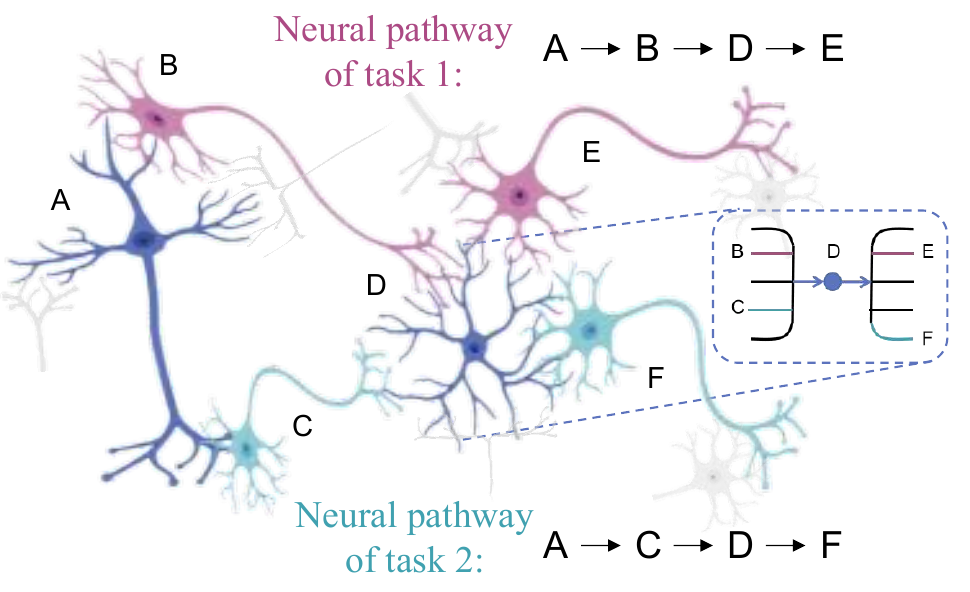}
	\caption{\textbf{Sparse neural pathways adaptively reorganize and collaborate for continual learning.} Purple neurons and cyan neurons are individual neurons for task 1 and task 2, respectively, and blue neurons are shared for both tasks. In the blue box, the different synapses of neuron D are utilized for different tasks and form sparse connections.}
	\label{fig0}
\end{figure}

The dynamic routing mechanism serves as a fundamental supporting framework for cognitive functions in the brain~\cite{voytek2015dynamic}. The brain adapts to varying cognitive demands through self-organized dynamic neural circuits~\cite{phillips2010dynamic,tau2010normal}, a process that relies on competition-cooperation interactions among neurons and neural pathways to enable efficient information routing~\cite{xerri2012plasticity}. This reorganization is jointly regulated by genetic factors and environmental inputs~\cite{huttenlocher2009neural}, while feedback mechanisms~\cite{kourtzi2006learning}, synchronized neural oscillations~\cite{grossberg1991synchronized}, and neuromodulatory systems~\cite{lipton1989neurotransmitter} work in concert to shape task-specific sparse neural circuits~\cite{jacobs1991reshaping,barron2021neural} via temporary synaptic silencing(Fig. \ref{fig0}). Even though the number of neurons no longer increases in adulthood~\cite{huttenlocher1990morphometric}, dynamic routing reorganization continues to support lifelong learning capabilities, ranging from basic sensory processing to higher-order social cognition.

Although existing continuous learning algorithms for artificial neural networks propose some solutions inspired by brain mechanisms, most of them are based on fixed dense network structures or continuously scaling network sizes, limiting their ability to adaptively switch and knowledge transfer between tasks as well as their energy efficiency. In addition, the vast majority of existing continuous learning algorithms are based on deep neural networks (DNNs), with little exploration of spiking neural networks (SNNs). Therefore, we employ centrally organized regulatory networks to adaptively reorganize sparse neural pathways according to task features in limited SNNs, which empowers bi-directional task-to-task knowledge transfer and adaptive functional repair capabilities.

The DNNs-based Continual Learning has been proposed mainly inspired by two categories of brain mechanisms: synaptic plasticity and structural plasticity.
Synapses are carriers of memory \cite{mayford2012synapses}, and synapse-based continuous learning algorithms can be divided into synaptic importance measure and knowledge distillation. The former restricts the plasticity of important synapses~\cite{kirkpatrick2017overcoming,zenke2017continual,aljundi2018memory}, and the latter uses ``soft-supervised'' information from old tasks to overcome catastrophic forgetting~\cite{li2017learning,murata2022learning}. Some recently proposed dual-network continuous learning algorithms~\cite{von2019continual,chandra2023continual,beaulieu2020learning,pham2020contextual} use the additional network to generate the main network weights or weight regulation coefficients. However, these algorithms use fixed dense networks in all tasks lacking sufficient memory capacity and sparsity to learn large-scale tasks. 

Inspired by the dynamic structural plasticity mechanisms of the brain, additional algorithms extend new network structures for new tasks~\cite{rusu2016progressive,yoon2017lifelong}, resulting in network consumption skyrocketing with the number of tasks, and being inconsistent with the energy efficiency of the brain. To solve this problem, the proposed subnetwork selection algorithms select a sparse subnetwork structure for each task by evolution~\cite{fernando2017pathnet}, pruning~\cite{dekhovich2023continual,golkar2019continual}, or reinforcement learning~\cite{xu2018reinforced,gao2022efficient}. These algorithms reduce energy consumption, but the subnetworks for all tasks are chosen from the initial network, hindering knowledge transfer between tasks.

Spiking neural networks (SNNs), as the third-generation artificial neural network ~\cite{Maass1997Networks}, simulate the discrete spiking information transfer mechanism in the brain ~\cite{zhao2022nature}. The basic unit spiking neuron integrates rich spatio-temporal information, which is more biologically plausible and energy efficient.
As a brain-inspired cognitive computing platform, SNNs have achieved comparable performance to DNNs in classification~\cite{wu2019direct,wu2018spatio}, reinforcement learning~\cite{patel2019improved,sun2023multi} and social cognition~\cite{zhao2023brain} modelling. Among the few existing SNNs-related continuous learning algorithms, HMN~\cite{zhao2022framework} uses DNNs to regulate the spiking thresholds of the neurons in SNNs. ASPs~\cite{panda2017asp} rely on spiking time-dependent synaptic plasticity (STDP) to overcome catastrophic forgetting. However, both of them are only suitable for shallow networks to accomplish simple tasks. DSD-SNN~\cite{han2023enhancing} introduces brain-inspired structural development and knowledge reuse mechanism that enable deep SNNs to accomplish continual learning and no longer need to save additional sub-network masks, but still suffers from the problem of ever-expanding network consumption. The contribution of SNNs in multi-task continual learning is still waiting to be further explored.

The human brain could dynamically reorganize neural circuits during continual development and learning processes. Especially, the adult brain possesses a nearly fixed number of neurons and connections~\cite{huttenlocher1990morphometric}, while it can incrementally learn and memorize new tasks by dynamically reorganizing the connections between neurons~\cite{bontempi1999time}.Inspired by this, we designed the adaptive dynamic routing reorganization Network (ADR-SNN) to spontaneously activate sparse neural pathways from a limited spiking neural network, thus enabling centrally regulated inter-task knowledge transfer and damage repair. 
 The main contributions of this paper can be summarized as follows:

\begin{itemize}
    \item [$\bullet$] Our proposed ADR-SNN model can adaptively reorganize to activate task-related sparse neural pathways without human intervention, reducing per-task energy consumption while enabling the limited SNNs to have a large number of sparse neural connectivity combinations, thus enhancing the memory capacity for more tasks. 
 
    \item [$\bullet$]
Extensive experiments demonstrate the superior performance of the proposed model on child-like simple-to-complex cognitive tasks and generalized datasets CIFADR100 and ImageNet.  In addition, the proposed model reveals outstanding strengths on more complex tasks, and knowledge backward transfer capability in which learning new tasks improves the performance of old tasks without replaying old task samples.
    
    \item [$\bullet$] The ADR-SNN model also shows the self-repairing ability to irreversible damage, in that when the structure of the SNN is pruned resulting in forgetting of acquired knowledge, the retained neurons and synapses can be automatically assigned to repair memory for the forgotten task without affecting other acquired tasks.
   
\end{itemize}

\section{Results}

\subsection{Spiking Neural Networks with adaptive dynamic routing reorganization Network (ADR-SNN) Framework}

\begin{figure*}[t]
	\centering 
	\includegraphics[width=0.98\linewidth]{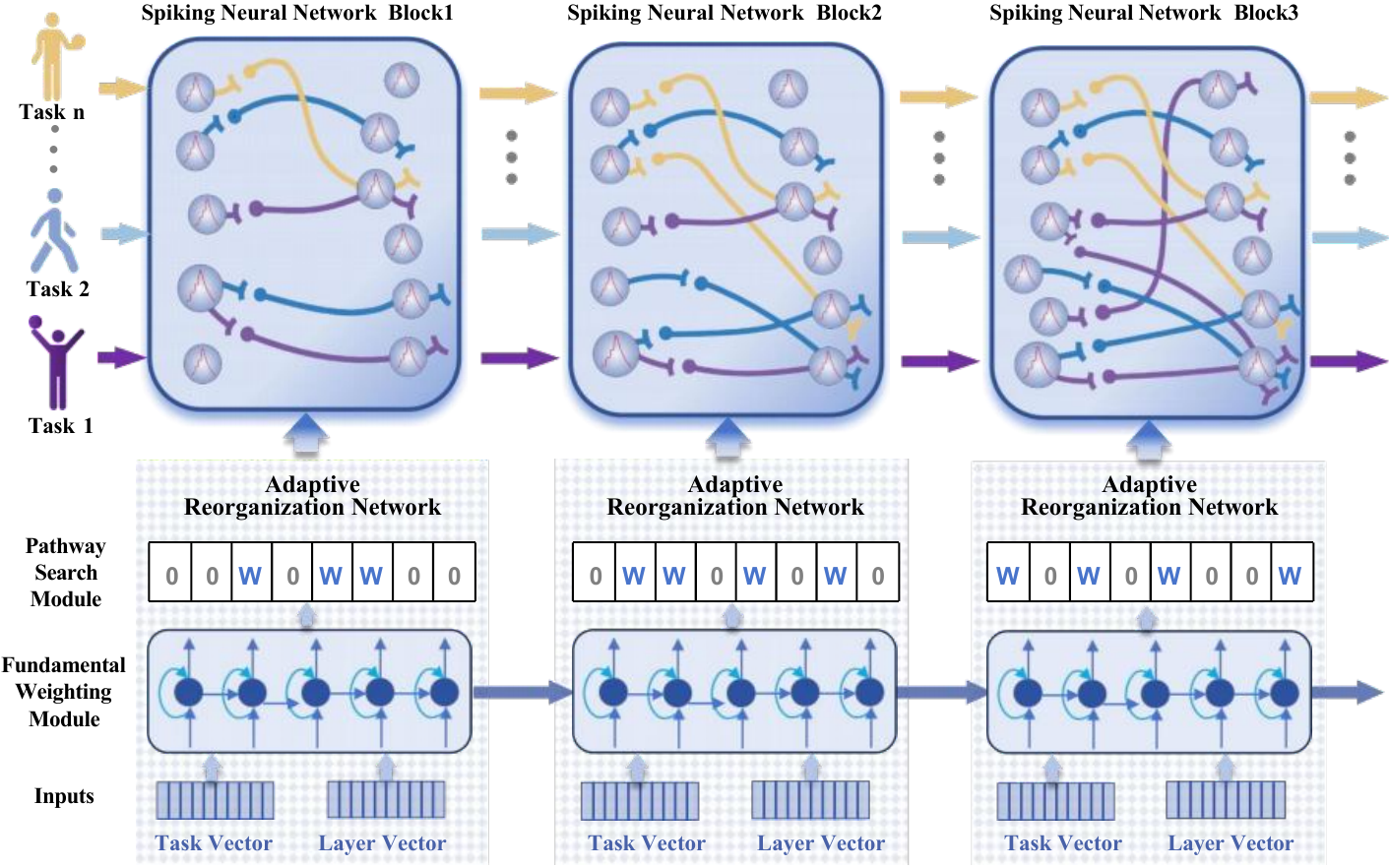}
	\caption{\textbf{The procedure of ADR-SNN model.} Each spiking neural network block in the proposed ADR-SNN model involves a adaptive dynamic routing reorganization Network (ADR-SNN)s which is responsible for selectively activating task-specific sparse pathways in the SNN. For example: the purple connections form the pathway for task 1. In particular, the adaptive dynamic routing reorganization Network (ADR-SNN) contains the fundamental weighing module and the path searching module. The large number of different combinations of connections allows the limited SNN to have the capacity to incrementally learn more n tasks.}
	\label{method fig}
\end{figure*}

The proposed ADR-SNN algorithm consists of two components: the SNN master network is responsible for accomplishing multi-task continual learning; the adaptive dynamic routing reorganization Network network ADR selects sparse subneural pathways of the SNN master network for each task. Each region of the SNN possesses an ADR networks, and the regions with multiple layers are divided according to the structure of the SNN (e.g. each block in a ResNet model represents a region) as Fig. \ref{method fig}. The SNN invokes different sparse neural circuits sequentially to accomplish different tasks under the regulation of the adaptive dynamic routing reorganization Network (ADR-SNN). 

The main network SNN uses commonly used LIF neurons with surrogate gradients to form the typical Resnet structure. The ADR networks' inputs are learnable task-related vector $x_t$ and layer-related vector $x_l$, as well as the state of the ADR network in the previous layer, and the outputs are the sub-neural pathways selected for the SNN in task $t$. The organizational regulatory network inputs sequentially pass through its own pathway search module and fundamental weighting module: the former selects 0/1 sparse neural pathways $P_t$ for the SNN, and the latter generates the basic weight strengths $W_t$ of the SNN, which together adaptively reorganize rich sparse sub-nets $P_t W_t$ for successive completion of different tasks. 

During the training stage, the ADR-SNN algorithm learns the ADR network's weights and learnable inputs end-to-end to better determine the sub-networks of the main SNN in different tasks, To overcome catastrophic forgetting, the reorganization network makes the neural pathways $P_t$ activated for each task as orthogonal as possible to minimize inter-task interference. Orthogonal activation pathways allow finite SNNs with a large number of different connection combinations to accomplish more tasks and achieve higher memory capacity. At the same time, we make the synaptic weights $W_t$ as equal as possible in each task to preserve the memory. In the testing stage, ADR network reorganizes the task-specific sparse structure and corresponding weights of the main SNN based on the inputs of the task $x_t$ and the layer $x_l$. Then, the selected sub-net in the main SNN accepts samples from the current task $t$ to complete the classification task. 

 Based on the flexible regulation of the ADR network, the main SNN could generate diverse sub-neural pathways, showing the potential of expanding the memory capacity. In this paper, we verify the performance, energy consumption, memory capacity, and backward transfer ability on child-like simple-to-complex multi-task learning, generalized continual learning datasets (CIFADR100 and ImageNet). More importantly, the separate ADR network shows high adaptability to structural mutations of the main SNN network.

\begin{figure*}[t]
	\centering 
	\includegraphics[width=1\linewidth]{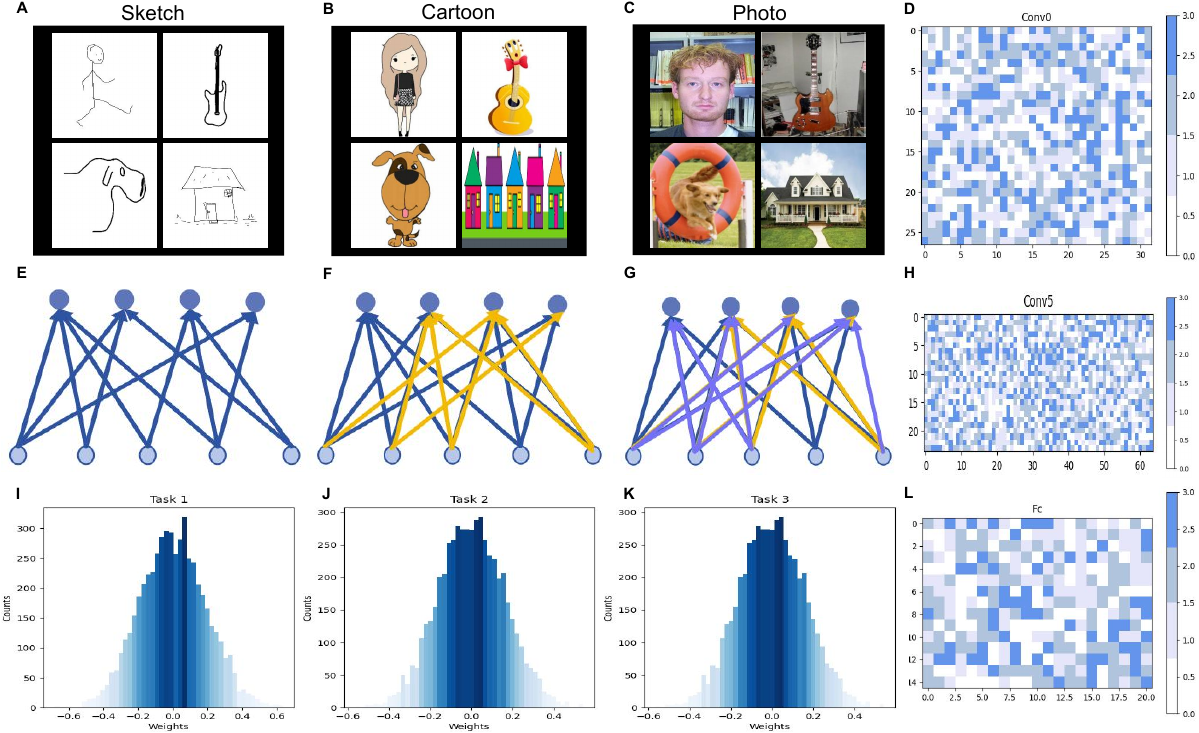}
	\caption{\textbf{Validation of child-like simple-to-complex continual learning. }\textbf{(A-C)} The simple to complex cognitive tasks include sketches, cartoons and photos. \textbf{(E-G)} Task-specific sparse pathways, for example, the blue, yellow and purple arrows represent the pathways for Task 1, Task 2 and Task 3 respectively in the fully connected output layer. \textbf{(D,H,L)} Visualization of synaptic activation counts in partial convolutional and fully connected layers. \textbf{(I-K)} Distribution of real-valued weights in the fully connected layer for three different tasks. }
	\label{res2 fig}
\end{figure*}

\subsection{Continual learning child-like simple-to-complex tasks}
During child growth and development, the brain gradually learns and memorizes hundreds of cognitive tasks. This process does not happen all at once, but starts with simple tasks and gradually progresses to more complex ones. In this paper, we simulate the developing process of child simple-to-complex cognition by sequentially learning sketches (3,929 images), cartoon drawings (3,929 images), and real photographs (1,670 images)~\cite{zhou2020deep}, as shown in Fig. \ref{res2 fig} A-C. Specifically, our SNN structure is ResNet18 and the reorganization network employs the LSTM with 96 hidden layer neurons.

During the simple-to-complex experiment, samples of three cognitive tasks are sequentially fed into the ADR-SNN model to learn. We monitored the SNN weights and task-specific pathways under the guidance of the adaptive dynamic routing reorganization Network (ADR-SNN)s during the learning process. We visualized the activation of partial weights in the fully connected output layer of the ResNet18 SNN for each of cognitive task, as shown in Fig. \ref{res2 fig} E-G. In the whole network, the adaptive dynamic routing reorganization Network (ADR-SNN)s activates different combinations of weights for different tasks, forming task-specific sparse neural pathways. This gives a single network the ability to accomplish multiple cognitive tasks while reducing mutual interference between tasks and saving the energy consumption required per task. 

Meanwhile, each synapse in our SNN was involved in a different number of task-specific pathways as shown in Fig. \ref{res2 fig} D,H,L. Some synapses were activated in all three task-specific pathways, some synapses were activated in only one task-specific pathway, while others constantly remained inactive. This suggests that in our ADR-SNN model, the pathways of different tasks share connections for processing common features as well as have their own unique connections for recognizing task-specific features. Moreover, there is a portion of connections in the network that remain inactive all the time, indicating that the memory space of the network still has the capacity to accomplish more tasks.

In addition, we statistically calculated the distribution of the real-valued weights of SNNs from the output of the fundamental weighting module in the ADR network for each cognitive task as shown in Fig. \ref{res2 fig} I-K. The results show that the distribution of the synaptic weights of the SNNs in different tasks has minor variations but generally stays the same. Unlike existing artificial neural networks in which past learned knowledge is lost due to large changes in weights, our ADR-SNN model applies similar weights in different tasks that can effectively avoid catastrophic forgetting. 

\subsection{Superiority in performance, energy consumption, memory capacity and backward transfer}
To verify the effectiveness of our ADR-SNN, we conduct experiments on child-like simple-to-complex continual learning tasks, and two generalized CIFADR100 and ImageNet datesets. For comparison, we replicate other continual learning algorithms to SNN, including: the EWC~\cite{kirkpatrick2017overcoming} and MAS~\cite{aljundi2018memory}, methods based on the modification of synaptic plasticity in a single network; HNET~\cite{von2019continual},  LSTM$\_$NET~\cite{chandra2023continual} and the DualNet~\cite{pham2021dualnet} methods based on the dual-network generating weights or weight regulation coefficients. In addition, we compare the rare previous SNN-based Continual Learning method DSD-SNN~\cite{han2023enhancing}, which belongs to the sparse structural extension methods. All of these methods were conducted in multiple times using the same LIF neurons and the same generational gradient training method in the SNN. 

\begin{figure*}[t]
	\centering 
	\includegraphics[width=1\linewidth]{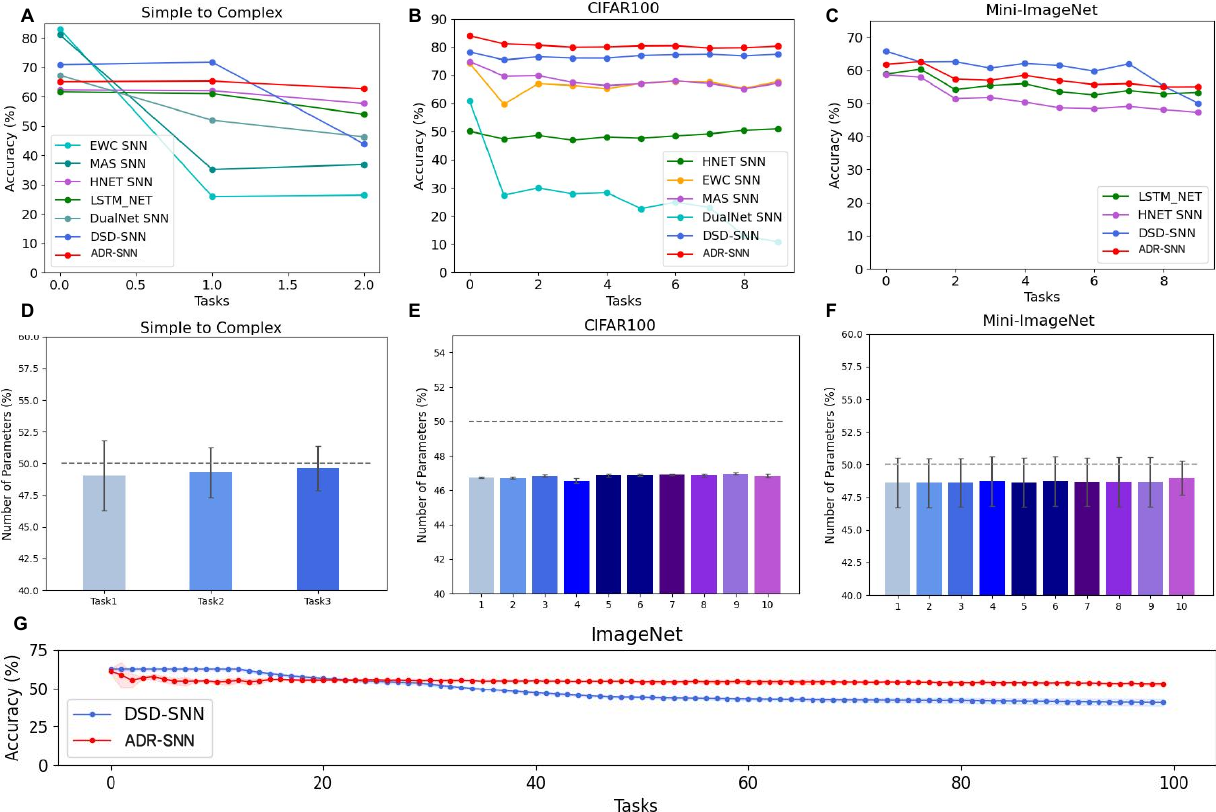}
	\caption{\textbf{The comparative performance of ADR-SNN on diverse continual learning tasks.} The average accuracy \textbf{(A-C)} and the number of inactive parameters \textbf{(D-F)} of the network for simple to complex cognitive tasks, the CIFADR100 and Mini-ImageNet datasets.  The average accuracy on the large scale ImageNet dataset \textbf{(G)}.}
	\label{res3 fig}
\end{figure*}

\subsubsection{Continual learning accuracy}
Fig. \ref{res3 fig} A-C,G is the average accuracy of task t and its previous tasks. We can find that our ADR-SNN algorithm maintains superior accuracy in all multi-task sequential learning of the simple-to-complex cognitive task, the CIFADR100, mini-ImageNet, and ImageNet datasets. For simple-to-complex cognitive task, our ADR-SNN achieves an average accuracy of 62.72$\pm$1.25\%, which is consistently higher than HNET SNN and LSTM$\_$NET SNN continual learning algorithms. 
Although, the EWC SNN and MAS SNN achieved high accuracy in the first simplest task, their learning and memorization abilities significantly decreased in the more complex tasks. The DSD-SNN achieves higher accuracy in the first two tasks than our ADR-SNN, but after adding the most complex tasks, the average accuracy of our ADR-SNN is higher than the DSD-SNN. This suggests that our ADR-SNN is able to gradually enhance the learning ability in the process of knowledge accumulation and better accomplish more complex tasks compared to other algorithms.

For CIFADR100, we tested continual learning scenarios with 5steps (each task contains 20 classes), 10steps (each task contains 10 classes), and 20steps (each task contains 5 classes). The accuracy comparisons are depicted in Tab. \ref{cifar}. For 10steps, our ADR-SNN achieves an average accuracy of 80.12$\pm$0.25\%, both consistently higher than other methods based on SNN or replicated in SNN. Compared to DSD-SNN, which has the second highest average accuracy and is the only known method implementing SNNs continual learning in CIFADR100, our ADR-SNN achieves a 2.20\% accuracy improvement. Besides, our ADR-SNN has accuracy rates of 73.48\%$\pm$0.46 and 86.65\%$\pm$0.20 in 5steps and 20steps, with 9.04\% and 5.48\% improvement than the second highest accuracy, respectively. In particular, in 20steps learning, the performance of the proposed model not only does not degrade with the number of tasks, but also achieves a higher accuracy, demonstrating the strong memory capability of our model. Compared to DNN-based continual learning algorithms, our model achieves superior accuracy with low energy consumption as shown in Tab. \ref{cifar2}. 

\begin{table}[htbp]
  \centering
  \resizebox{4.8in}{!}{
        \begin{tabular}{l|r|r|r|r|r|r}
    \toprule
    \multicolumn{1}{c|}{\multirow{2}[4]{*}{Method }} & \multicolumn{2}{c|}{5steps} & \multicolumn{2}{c|}{10steps} & \multicolumn{2}{c}{20steps} \\
\cmidrule{2-7}      & \multicolumn{1}{c|}{Acc (\%) } & \multicolumn{1}{c|}{Std (\%)} & \multicolumn{1}{c|}{Acc (\%)} & \multicolumn{1}{c|}{Std (\%)} & \multicolumn{1}{c|}{Acc (\%)} & \multicolumn{1}{c}{Std (\%)} \\
    \midrule
    EWC~\cite{kirkpatrick2017overcoming} & \multicolumn{1}{r}{45.89} & \multicolumn{1}{r}{0.96} & \multicolumn{1}{r}{61.11} & \multicolumn{1}{r}{1.43} & \multicolumn{1}{r}{50.04} & 4.26 \\
    SI~\cite{zenke2017continual} & \multicolumn{1}{r}{66.92} & \multicolumn{1}{r}{0.17} & \multicolumn{1}{r}{64.81} & \multicolumn{1}{r}{1.00} & \multicolumn{1}{r}{61.10} & 0.82 \\
    MAS~\cite{aljundi2018memory} & \multicolumn{1}{r}{61.88} & \multicolumn{1}{r}{0.27} & \multicolumn{1}{r}{64.77} & \multicolumn{1}{r}{0.78} & \multicolumn{1}{r}{60.40} & 1.74 \\
    HNET~\cite{von2019continual}  & \multicolumn{1}{r}{48.69} & \multicolumn{1}{r}{0.37} & \multicolumn{1}{r}{63.57} & \multicolumn{1}{r}{1.03} & \multicolumn{1}{r}{70.48} & 0.25 \\
    LSTM\_NET~\cite{chandra2023continual}   & \multicolumn{1}{r}{60.11} & \multicolumn{1}{r}{0.88} &  \multicolumn{1}{r}{66.61}  & \multicolumn{1}{r}{3.77}   &  \multicolumn{1}{r}{79.96}  & \multicolumn{1}{r}{0.26}  \\
    DSD-SNN~\cite{han2023enhancing} & \multicolumn{1}{r}{64.44} & \multicolumn{1}{r}{0.24} & \multicolumn{1}{r}{77.92} & \multicolumn{1}{r}{0.29} & \multicolumn{1}{r}{81.17} & 0.73 \\
    \textbf{Our ADR-SNN} & \multicolumn{1}{r}{\textbf{73.48}} & \multicolumn{1}{r}{\textbf{0.46}} & \multicolumn{1}{r}{\textbf{80.12}} & \multicolumn{1}{r}{\textbf{0.25}} & \multicolumn{1}{r}{\textbf{86.65}} & \textbf{0.20} \\
    \bottomrule
    \end{tabular}}%
  \caption{Accuracy comparisons on 5steps, 10steps and 20steps for CIFADR100.}
  \label{cifar}%
\end{table}%

In the ImageNet dataset, which has a larger sample scale and a larger number of sample classes, we first randomly selected 100 classes to form the Mini-ImageNet dataset and divided them into ten tasks. As shown in Fig. \ref{res3 fig}C, our ADR-SNN model achieves consistent superiority over HNET SNN and LSTM$\_$NET SNN continual learning algorithms (EWC SNN and MAS SNN fail in the Mini-ImageNet). Although the DSD-SNN outperforms our ADR-SNN in the earlier tasks, its accuracy gradually decreases during learning the subsequent tasks. Eventually, our ADR-SNN achieves a higher average accuracy than DSD-SNN in all tasks. This demonstrates that our model has more memory capacity and is able to learn and memorize more tasks. In contrast, DSD-SNN relies on energy-consuming structure growth, which is not competent for learning more tasks (although it brings improvement on the first few tasks).

In addition, we evaluated our ADR-SNN on the complete ImageNet dataset, which is the first time the SNN-based model achieves continual learning on the large-scale ImageNet dataset. Compared to the competitive DSD-SNN method, our model consistently achieves superior performance after learning 20 tasks. The accuracy rate of our ADR-SNN remains essentially stable over 100 tasks of the continual learning. This indicates that our ADR-SNN has a large memory capacity to constantly learn new tasks.

\subsubsection{Energy consumption }
Under the effect of the adaptive dynamic routing reorganization Network (ADR-SNN)s, our model activates a task-specific sparse pathway per task and uses only a portion of the network parameters, reducing the energy consumption of each task. Fig. \ref{res3 fig} D-F depicts the number of synapses used for each task in the three datasets respectively. The activation rate of the main SNN  is only between 40\% and 50\%. The average compression rate of our ADR-SNN is 52.37\%,53.88\% and 53.08\% in the simple-to-complex cognitive task, CIFADR100 and Mini-ImageNet datasets, respectively. 

Additionally, we compare the network energy consumption of the proposed methods in terms of the number of connections, neurons, floating point operations per second (FLOPs), and computational energy, as shown in Tab. \mbox{\ref{tab3}}. The computational energy is calculated following the widely adopted approach from ~\mbox{\cite{chakraborty2021fully}}, specifically for DNNs.
\begin{equation}
	\label{es}
	E_{SNN}=FLOPS_{SNN}*E_{MAC}
\end{equation}
where $E_{MAC}=4.6pJ$ is the energy consumption of multiply-accumulate (MAC) operations. For SNN: 
\begin{equation}
	\label{es}
	E_{SNN}=FLOPS_{SNN}*E_{AC}*T
\end{equation}
where $E_{AC}=0.9pJ$ is the energy consumption of accumulate (AC) operations.

The results show that our ADR-SNN algorithm significantly reduces the energy consumption, with the network connection count being only 1.6$ \times10^6$. This is because, compared to synaptic regularization~\cite{kirkpatrick2017overcoming,douillard2020podnet} and sample replay ~\cite{rebuffi2017icarl} methods that rely on fixed and dense structures, ADR-SNN only uses a sparse subset of connections for each task. In contrast to structure expansion methods~\cite{han2023enhancing,yan2021dynamically}, our overall network structure does not expand with the increasing number of ttasks. Instead, it reorganizes the existing structure to select new sub-networks.

Compared to traditional DNN methods, our SNN-based approach significantly reduces computational energy by replacing the multiply-accumulate operation with the accumulate operation. With the same FLOPs, SNN-based methods require less computational energy. Due to the reduced FLOPs and the inherent efficiency of SNNs, our method achieves the lowest computational energy of 1.1$ \times10^9$pJ. In summary, the combination of the sparsity introduced by the proposed ADR method and the high efficiency of SNNs enables our ADR-SNN approach to significantly reduce network energy consumption.

\begin{table*}[t]
    \centering
    \caption{Energy consumption comparisons for 10steps CIFADR100.}
    \resizebox{6in}{!}{
\begin{tabular}{lccccc}
    \toprule
    \toprule
    \multicolumn{1}{c}{\textbf{Method}} & \makecell{\textbf{Memory} \\ \textbf{ Method}} &  \makecell{\textbf{Number of} \\ \textbf{Connections}} & \makecell{\textbf{Number of} \\ \textbf{ Neurons}} & \textbf{FLOPs} & \makecell{\textbf{Computational} \\ \textbf{ Energy}} \\
    \midrule
    EWC SNN~\cite{kirkpatrick2017overcoming} & Regularization & 11.2$ \times10^6$ & 3840 & 11.1$ \times10^8$ & 4.0$ \times10^9$ pJ \\
    MAC SNN~\cite{aljundi2018memory} & Regularization & 11.2$ \times10^6$ & 3840 & 11.1$ \times10^8$ & 4.0$ \times10^9$ pJ \\
    PODNet DNN~\cite{douillard2020podnet} & Regularization & 11.2$ \times10^6$ & 3840 & 11.1$ \times10^8$ & 5.1$ \times10^9$ pJ \\
    iCaRL DNN~\cite{rebuffi2017icarl} & Replay & 11.2$ \times10^6$ & 3840 & 11.1$ \times10^8$ & 5.1$ \times10^9$ pJ \\
    DER++ DNN~\cite{buzzega2020dark} & Replay & 11.2$ \times10^6$ & 3840 & 11.1$ \times10^8$ & 5.1$ \times10^9$ pJ \\
    DSD-SNN~\cite{han2023enhancing} & Expansion & 115.8$ \times10^6$ &  5362 & 38.4$ \times10^8$  & 13.8$ \times10^9$ pJ \\
    DER DNN~\cite{yan2021dynamically} & Expansion & 61.6$ \times10^6$ & 21472 & 61.1$ \times10^8$ & 28.1$ \times10^9$ pJ \\
    SCA-SNN~\cite{han2024similarity} & Expansion & 8.4$ \times10^6$ & 3111 & 9.2$ \times10^8$ & 3.3$ \times10^9$ pJ \\
     \textbf{Our ADR-SNN}  &  Reorganization  & 1.6$ \times10^6$ & 1572 & 1.5$ \times10^8$& 1.1$ \times10^9$ pJ\\
    \bottomrule
    \bottomrule
    \end{tabular}}%
		\label{tab3}%
	\end{table*}%
    
\subsubsection{Backward transfer capability}
In the brain, not only has the forward transfer capacity that the past learned knowledge can assist the learning of new tasks, but also the learning of new tasks will improve the performance of the past tasks. In recent years, the forward transfer capability of continual learning algorithms has been proposed and validated in many models, but currently there are fewer models with backward transfer capability. This is due to the fact that many models freeze the connections related to previous tasks from being updated again or reduce their weight plasticity~\cite{wortsman2020supermasks,yoon2017lifelong,han2023enhancing}, hindering the backward transfer between tasks. As for our ADR-SNN model, it retains the ability of past task-related synapses to be fine-tuned according to the new task while selecting new pathways for the new tasks. 

\begin{figure*}[t]
	\centering 
	\includegraphics[width=1\linewidth]{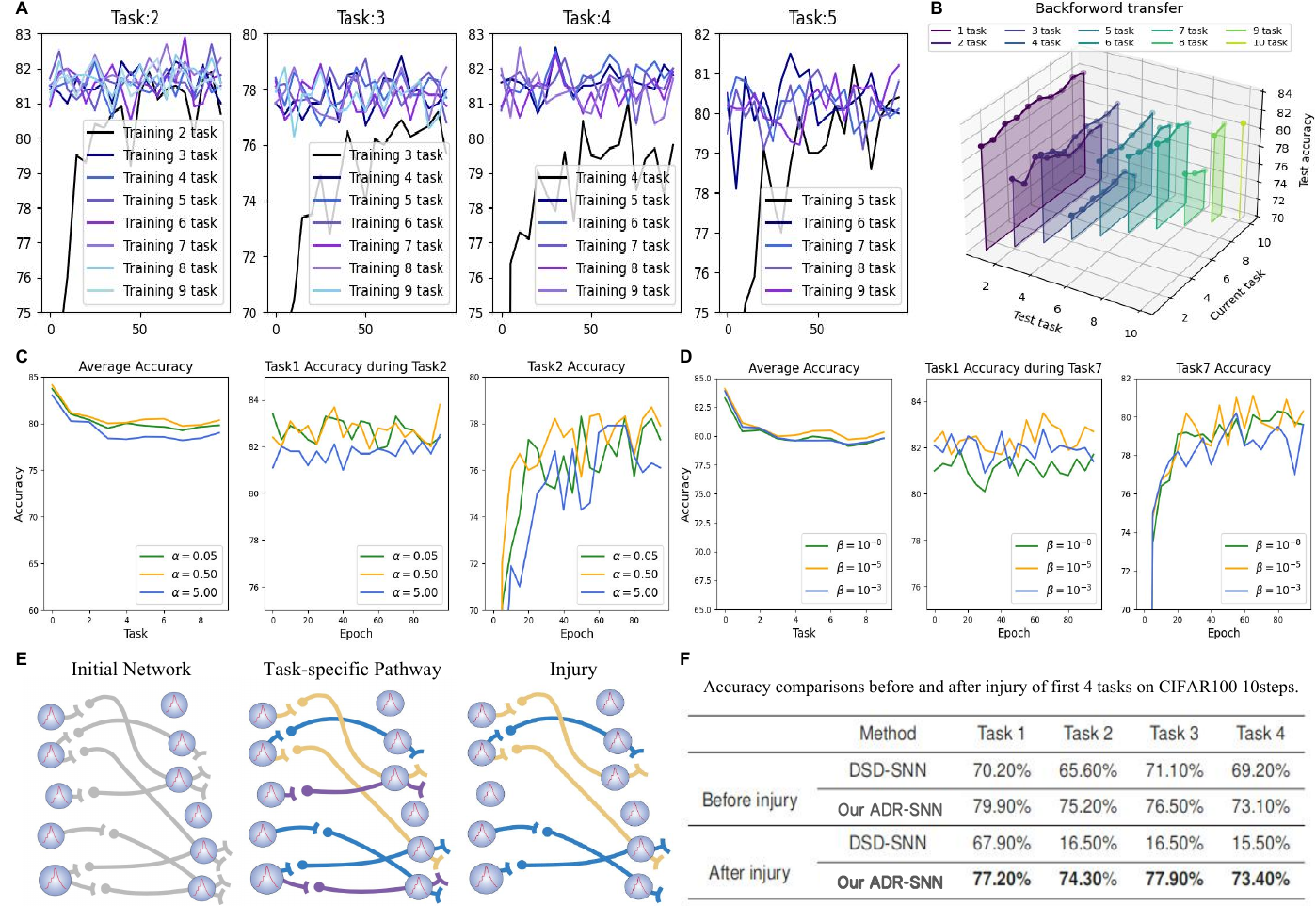}
	\caption{\textbf{(A-B)} The current test accuracy of past learned tasks in our ADR-SNN model. \textbf{(C-D)} The effect of memory loss coefficient and orthogonal loss coefficient on performance. \textbf{(E)} Injury schematic, containing the initial network, the network with task-specific pathways assigned and the network after injury task 1 of SNNs. \textbf{(F)} Accuracy comparisons before and after injury of first four tasks on CIFADR100 10steps.}
	\label{res4 fig}
\end{figure*}

In order to verify the backward transfer capability of the proposed method, we analyze the performance changes during the learning process of all tasks. As shown in Fig. \ref{res4 fig}A, taking the learning of task 2 as an example, the accuracy of the second task after the third task is learned (darkblue line) is higher than when the second task was first learned (black line), and the similar phenomenon occurs in the learning of other tasks. From the whole learning process as shown in Fig. \ref{res4 fig}B, the final test accuracy of each previous task is mostly positively sloping with the learning of the current task, indicating that the performance of the previous task is in an integrally upward trend with the increase of the learning task. In addition with the learning of subsequent tasks, the accuracy stability of the previous task also increases to some extent. This suggests that the knowledge learned from later tasks in our ADR-SNN model could contribute to a better completion of the previous task without retraining the previous task, realizing the backward transfer of knowledge.

\subsubsection{Ablation studies}

The challenge for neural networks to maintain a stable memory of acquired knowledge while possessing plasticity to learn new tasks is called the "stability-plasticity" dilemma, which is the core problem of Continual Learning. In our ADR-SNN model, we maintain stable memory by making the real weights of different tasks as equal as possible through memory loss, and enable the network to learn more new tasks by making the pathway connections of different tasks as different as possible through orthogonality loss. The $\alpha$ and $\beta$ respectively control the contribution of memory loss and orthogonality loss in network optimization. We analyze the effect of different $\alpha$ and $\beta$ parameters on the stability and plasticity of the proposed model. As shown in Fig. \ref{res4 fig}C, when $\alpha$ is too large, the network forces the real weights of two different tasks to be too identical, resulting in a drop in the accuracy of both the old and new tasks. For example, during the learning process of task 2, when $\alpha=5$, the accuracy rate of task 2 is inferior, while the accuracy for the previously learned task 1 is also the lowest. For CIFADR100, when $\alpha$ is less than 0.5, the network maintains the memory of the old tasks without affecting the learning of the new tasks. Among them, when $\alpha$ = 0.5, the highest average accuracy rate was consistently observed over the 10 tasks learning process.

For the orthogonal loss coefficient $\beta$, the SNN achieves the best performance of 80.12\% when $\beta=10^{-5}$ as shown in Fig. \ref{res4 fig}D. When $\beta$ is minor, taking the learning process of task 7 as an example, the new task 7 has a higher accuracy, but the test accuracy of the old task 1 is lower. This is because if $\beta$ is smaller, the orthogonal loss is smaller and the contribution of the classification loss is larger, so the new task performs better. However, this simultaneously results in a higher overlap between task-specific pathways, where the knowledge learned with the old task is easily interfered and disturbed. Conversely, when $\beta$ is larger, the contribution of orthogonality loss is greater and the pathways of the old and new tasks are more orthogonal. As a result, there is stronger memorization and a higher test accuracy for the old task 1. Meanwhile, due to the overemphasis on pathways orthogonality, the accuracy on the new task 7 slightly drops. As shown in Fig. \ref{res4 fig}B-C, our ADR-SNN algorithm achieves an acceptable average accuracy for different loss parameters. This suggests that our model is robust and stable.

\subsection{The injury self-repair capability of ADR-SNN}
The adaptive reorganization in the brain is also reflected in the self-repairing capability after injury~\cite{dresp2020seven}. Our brain is a robust system that can withstand various perturbations or even multiple micro-strokes without significant deleterious impact~\cite{joyce2013human}. After the injury of one subsystem, the brain is able to adaptively reorganize and select new subsystems (synapses or networks) to perform functions previously handled by the injured subsystem, which is known as the functional plasticity~\cite{freed1985promoting,anderson2005functional}. For example, in the visual system of cats, after parts of the retina are lesioned and injured, up to 98\% of afferent neurons generate new receptive fields in the residual uninjured areas by optimizing neuronal connectivity when receiving previous inputs again ~\cite{sabel2011vision}. Similarly, after partial peripheral nerve injury in the monkey brain, connections from the somatic surface to the primary senory cortex undergo substantial reorganization. Electrophysiological recordings show that stimuli inputs that are previously responded to in the injured area now produce a reaction in the neighbouring uninjured area~\cite{xerri1998plasticity}. This self-organizing repair capacity is also found extensively in the medial prefrontal cortex, hippocampus, and amygdala~\cite{bremner2007structural}.

To simulate this brain mechanism and verify the self-repairing ability of our model, we pruned part of the connections that are unique to task 1 after continuously learning the first 4 tasks as shown in Fig. \ref{res4 fig}E. The pruned connections are no longer used for all tasks. When our model received samples of task 1 again, the ADR-SNN was able to self-organize to select new neural pathways among the remaining available synapses without affecting the performance of the previously learned other tasks 2, 3 and 4. The experimental results show that the accuracy of our model in task 1 kept stable pre and post-injury, and under the condition of not replaying the other tasks, the tasks 2, 3 and 4 maintained their original performance and were not significantly affected (and even improved slightly), as shown in Fig. \ref{res4 fig}F. Before and after the injury, our model slightly decreased by 0.9\% for task 2 and improved by 1.4\% and 0.3\% for tasks 3 and 4, respectively, whereas the DSD-SNN average accuracy of tasks 2, 3 and 4 dropped dramatically from 68.60\% pre-injury to 16.12\% experienced catastrophic forgetting. This is due to the fixed pathway of DSD-SNN for all tasks that lacked the ability of adaptive regulation. Overall, the experiments demonstrate that our reorganization enables a brain-inspired injury self-repair capability.

\section{Discussion}
During human lifelong learning, the nerve centre of the brain is self-organized to flexibly modulate neural circuits according to the characteristics of different tasks~\cite{guy2021human,zhang2001electrical,arain2013maturation}, selectively activating some of appropriate neuronal connections to efficiently complete incremental tasks. Neuroscience studies have shown that the neural network scale of the adult brain hardly changes anymore~\cite{huttenlocher1979synaptic}, but it can form temporary task-specific neural pathways by dynamically reorganizing existing neurons and synapses~\cite{hartwigsen2017rapid,takehara2003time}. The neural pathway is activated when needed, and during the rest of the time its elements are involved in forming neural pathways for other tasks~\cite{bontempi1999time}. Inspired by this, we propose a brain-inspired continual learning algorithm with spiking neural networks based on adaptive dynamic routing reorganization Network (ADR-SNN)s. Our model dynamically builds rich combinations of task-specific sparse neuronal connections for different tasks in limited SNNs, giving the SNNs the capacity to continually learn more complex tasks and more tasks.

Different from other continual learning algorithms for DNNs and SNNs, the proposed algorithm constructs sparse pathways that are self-organizingly selected by the reorganization network integrating the past learned knowledge and the current task features, rather than being artificially designed. Compared with existing synaptic regularization algorithms~\cite{li2017learning,cha2021co2l,murata2022learning} and neuroregulation algorithms~\cite{beaulieu2020learning,ben2022context,von2019continual,chandra2023continual}, our ADR-SNN uses only a portion of the neural connections for each task, reducing task energy consumption while alleviating forgetting due to the interference between tasks. For example, EWC~\cite{kirkpatrick2017overcoming} and MAX~\cite{aljundi2018memory} use all the synaptic connections of the 6.9$\times 10^5$ parametric network, while our ADR-SNN uses an average of only 46.12\% of the synaptic connections with 3.2$\times 10^5$ parameters for each task, achieving 80.12\% $\pm$ 0.25\% accuracy that is 19.01\% and 15.35\% higher than EWC and MAX, respectively. Meanwhile, compared with classical structure expansion algorithms~\cite{yan2021dynamically,hung2019compacting,siddiqui2021progressive}, our ADR-SNN is able to fully utilize the limited network to form rich connection combinations without having to grow new neurons for the new task; and temporarily inhibit some neural connections for subsequent tasks instead of actually pruning irrelevant connections for the current task. In addition, unlike the common operation in structure expansion algorithms that freezes the neurons and synapses related to the old task~\cite{wortsman2020supermasks,yoon2017lifelong,han2023enhancing}, each of our neurons and synapses is continuously learned and optimized, and thus the proposed algorithm achieves the knowledge backward transfer capability that the learning of the new task improves the performance of the old task without replaying the old task samples.

To validate the effectiveness of the proposed ADR-SNN, we conducted extensive experiments on different tasks and datasets. The results in child-like simple-to-complex cognitive task indicate that our ADR-SNN achieves higher performance than other methods in more complex tasks and realizes the highest average accuracy. Due to its ability to flexibly select sparse neural connections, when an already learned neural pathway is damaged and no longer usable, our model is able to adaptively select suitable alternative pathways from the remained network to repair the structure and function like the biological brain, and does not affect other tasks already learned. Furthermore, our ADR-SNN belongs to the pioneering exploration of SNN-based Continual Learning algorithms in the large-scale dataset ImageNet. Experiments in the generalized datasets CIFADR100 and ImageNet demonstrate that our ADR-SNN achieves superior performance in SNN-based Continual Learning algorithms. In summary, our ADR-SNN model continuously learns more tasks using less energy by self-organizingly constructing task-specific sparse neural pathways, which opens the path for building brain-inspired flexible and adaptive efficient Continual Learning.

\section*{Method}
The proposed ADR-SNN algorithm consists of two parts, one part is the main network SNN which sequentially performs multi-task sequential learning, and the different network pathways activated by SNNs in different tasks are determined by the other part is the adaptive dynamic routing reorganization Network (ADR-SNN). The computational details of the adaptive dynamic routing reorganization Network (ADR-SNN) and the main network SNN are as follows.

\subsection{SNN main network}
The main network receives the task input samples, and uses the activated sub neural pathways that the reorganization network produces to output the task classification prediction. Due to its bio-interpretability and energy-efficiency advantages, spiking neural network (SNN) is suitable as a basic platform for exploring brain-inspired multi-task sequential learning, and therefore we adopt SNN as the main network. 

SNN uses discrete spike sequences containing dynamic spatio-temporal dimensions. In the spatial dimension, the spiking neuron $i$ synthesizes the spike input $S_j$ from the presynaptic neuron $j$ to form the input current $I_i$; in the temporal dimension, the membrane potential $U_i$ of the spiking neuron accumulates the past spike information while receiving the current input current. In ADR-SNN, we used the common leaky integrate-and-fire 
 (LIF)~\cite{Lapicque1999L} spiking neuron with the following membrane potential $U_i$ and spike $S_i$ calculation formula:
 
\begin{equation}
    I_i^t=\sum_{j=1}^{M} P_t^{ij} S_j^t
    \label{I}
\end{equation}

\begin{equation}
    U_i^t=\tau U_i^{t-1}+I_i^t
    \label{U}
\end{equation}

\begin{equation}
	\label{S}
	S^t_{i}=\left\{\begin{matrix}
		1, &  U_i^t\geq V_{th}\\ 
		0, &  U_i^t < V_{th}
	\end{matrix}\right.
\end{equation}

where $\tau=0.2$ is the time constant and $t=4$ is the spiking time window. The spiking neuron receives the spike sequence input of length $t$ and calculates the membrane potential at time $t$ according to Eq. \ref{U}. When the membrane potential at time $t$ exceeds the spike firing threshold $V_{th}$, the neuron outputs 1 at time $t$; conversely the neuron outputs 0. Discrete spikes result in spiking neurons being non-microscopic. To solve this problem, we used the Qgradgate~\cite{qin2020forward} surrogate gradient method to approximate the gradient of the spike output as follows:
\begin{equation}
	\label{o}
    \frac{S_i^{t}}{U_i^{t}}=
        \begin{cases}
        0, & |U_i^{t}| > \frac{1}{\lambda} \\
        -\lambda ^2|U_i^{t}|+\lambda, & |U_i^{t}| \leq \frac{1}{\lambda}
        \end{cases}
\end{equation}
where constant $\lambda=2$. Above all, the discrete spatio-temporal spiking information transfer method reduces the energy consumption and enhances the knowledge representation capability of the spiking neural networks.

\subsection{adaptive dynamic routing reorganization Network (ADR-SNN)}
We use the same reorganization network to generate different sparse pathways $P_t$ and weight $W_t$ for different tasks $t$. An entire reorganization network consists of multiple sub-reorganization networks one-to-one responsible for each region of the main SNN. Each sub-reorganization network includes a fundamental weighting module and a pathway search module. 

To output task-specific pathways for different tasks with a single reorganization network, our reorganization network receives learnable task-relevant inputs $x_t$ and layer-relevant inputs $x_l$ as Eq. \ref{x}. During training, both of them adaptively learn representative information for different tasks and different layers, respectively; during testing, the learned $x_t$ and $x_l$ guide the reorganization network to output different specific sparse pathways. 
\begin{equation}
    x_{input}=\{x_t,x_l\}
    \label{x}
\end{equation}

\subsubsection{Fundamental Weighting Module} 
To synergize the synaptic activities between different regions and different layers, the recurrent neural network acts as the fundamental weighting module to output the real-valued weights $W_t$ prepared for the SNNs. In SNN, the weights of each layer are not independent, but work together to affect the overall performance. Thus we use the Long Short-Term Memory Network (LSTM) to synthesize information from previous layers and find the superior state of the current layer. Specifically, the hidden state of layer $l$ is the state of the previous layer $W_{t,l-1}$; in particular, the hidden state of the first layer of each region is the output of the last layer of the previous region. The fundamental weights $W_t$ are calculated as follows:

\begin{equation}
   W_{t,l}=LSTM(x_{input}, hidden)=LSTM(concat(x_t,x_l), W_{t,l-1})
    \label{w}
\end{equation}
where $concat$ denotes the connection of two vectors $x_t,x_l$, and $LSTM$ represents the standard LSTM network computational function.

\subsubsection{Pathway Search Module}
Based on the fundamental weights $W_{t,l}$, the pathway search module is responsible for deciding whether the activation or inhibition of each weight to self-organize the task-specific sparse neural pathways. Inspired by the differentiable structure search algorithm~\cite{liu2018darts,li2021differentiable}, each synapse has two states, active and inactive, respectively corresponding to a learnable synaptic selection parameter $A_{t,l}$ and $\widetilde{A_{t,l}}$. When the learnable parameter $A_{t,l}$ is greater than $\widetilde{A_{t,l}}$, activating the synapses is preferable for the performance of the current task compared to inhibiting it. That is, if $A_{t,l} > \widetilde{A_{t,l}}$, the synapse is activated. Otherwise, when the learnable parameter $\widetilde{A_{t,l}}$ is greater than $A_{t,l}$, the synapse s is supposed to be inhibited. Hence, the selection formula of sparse neural pathways $P_{t,l}$ for task $t$ is as follows:
\begin{equation}
	\label{p}
    P_{t,l}=
        \begin{cases}
        W_{t,l}, & A_{t,l} \geq \widetilde{A_{t,l}}\\
        0, & A_{t,l} < \widetilde{A_{t,l}}
        \end{cases}
\end{equation}
In order to adaptively decide the neural pathway according to the task, we convert the equation \ref{p} into a differentiable form as:
\begin{equation}
	\label{p2}
    P_{t,l}=W_{t,l} \cdot A_{t,l} \cdot Max(A_{t,l},\widetilde{A_{t,l}})
\end{equation}

\subsection{The plasticity - stability procedure of ADR-SNN}
During training, the SNN main network receives task input sample data $D_t$ and the neural pathway $P_{t,l}$ exported by the modulation network, and outputs the classification prediction $\hat{y}$. Then, we compute the training loss $L$, where the learnable parameters include the task-related inputs $x_t$, layer-related inputs $x_l$, synaptic selection parameter $A_{t,l}, \widetilde{A_{t,l}}$, and the LSTM weights.

For the balance of stability and plasticity, the loss function $L$ is divided into three parts: classification loss $L_{class}$, memory loss $L_{mem}$ and orthogonal loss $L_{orth}$. Firstly, the normal categorization loss $L_{class}$ aims to improve the learning performance of the current task. Our ADR-SNN model uses the cross-entropy loss: 

\begin{equation}
	\label{lc}
    L_{class}=\sum y \log{\hat{y}}=\sum y \log{(SNN(D_t, P_{t,l}))}
\end{equation}
Where y is the sample classification truth. 

Secondly, the memory loss $L_{mem}$ aims to make the real-valued weights $W_{t,l}$ as constant as possible in each task $t$ as Eq. \ref{mm}. Memory loss ensures that the fundamental weights $W_{t,l}$ do not change significantly across tasks to maintain the stability of learned knowledge. 

\begin{equation}
   L_{mem}=\Vert W_{t,l}-W_{t-1,l} \Vert_2
    \label{mm}
\end{equation}
And orthogonal loss $L_{orth}$ expects the sparse pathways $P_{t,l}$ selected for different tasks to be as different as possible to reduce interference between tasks, while making the SNN plastic to learn more new tasks through different connection combinations. The orthogonal loss is calculated as follows:

\begin{equation}
   L_{orth}=\sum_{k=1}^{t} P_{t,l} P_{k,l}
    \label{or}
\end{equation}
Moreover, we attempt to make $A_{t,l}$ and $\widetilde{A_{t,l}}$ close to 0.5 to stabilize the selection of task-specific sparse pathways. 

The combination of the latter two not only saves energy, but also allows the limited network to learn more tasks achieving larger memory capacity. The loss function $L$ is calculated as follows:

\begin{equation}
    L= L_{class}+\alpha L_{mem}+\beta L_{orth}
    \label{l}
\end{equation}
where $\alpha$ and $\beta$ are the constant coefficient.

In the testing process of our model, the reorganization network inputs task-relevant inputs $x_t$ and layer-relevant inputs $x_l$, self-organized outputs task-specific sparse pathways $P_{t,l}$; the SNN receives picture samples $D_t$ and task-specific pathways $P_{t,l}$, outputs prediction class $\hat{y}$. We present the specific procedure of our ADR-SNN algorithm as follow Algorithm:

 \begin{algorithm}[h]
	\caption{The ADR-SNN algorithm}
	\label{alg2}
	\KwIn{Dataset $D_t$ for each task $t$.}
    Initialization: randomly initialize learnable input $x_{t},x_{l}$ and learnable synaptic selection parameter $A_{t,l},\widetilde{A_{t,l}}$.\\
	\KwOut{Prediction Class $y$.}
	\For{$D_t$ in sequential task $t$}{
        \For{ $e$ in $Epoch$}{
            Calculating $W_{t,l}$ with learnable input $x_{t},x_{l}$ as Eq. \ref{x}-\ref{w};\\
            Selecting $P_{t,l}$ with synaptic parameter $A_{t,l},\widetilde{A_{t,l}}$ as Eq. \ref{p}-\ref{p2};\\
            SNN forward prediction $\hat{y}=SNN(D_t, P_{t,l})$ as Eq. \ref{I}-\ref{S};\\
            Calculating the training loss as Eq. \ref{lc}-\ref{l};\\
            Backpropagation to update the learnable parameter; 
		}
	}
\end{algorithm}	

\section*{Data availability}
The data used in this study are available in the following databases.

\noindent The simple-to-complex cognitive task data~\cite{zhou2020deep}: \\ \href{https://github.com/robertofranceschi/Domain-adaptation-on-PACS-dataset}{https://github.com/robertofranceschi/Domain-adaptation-on-PACS-dataset}. 

\noindent The CIFADR100 data~\cite{krizhevsky2009learning}: \\ \href{http://www.cs.toronto.edu/~kriz/cifar.html}{http://www.cs.toronto.edu/~kriz/cifar.html}. 

\noindent The ImageNet data\cite{deng2009imagenet}: \\ \href{https://image-net.org/}{https://image-net.org/}.

\section*{Acknowledgments}
	This work is supported by the Strategic Priority Research Program of the Chinese Academy of Sciences (Grant No. XDB1010302), the National Natural Science Foundation of China (Grant No. 62106261), Chinese Academy of Sciences (Grant No. ZDBS-LY-JSC013). We are specially grateful to Dr. Mu-ming Poo for his invaluable guidance and inspiration. His profound academic insights and unwavering support were instrumental in the successful completion of this study
 
\bibliographystyle{naturemag}
\bibliography{nature-communications}

\section*{Contributions}
B.H.,F.Z. and Y.Z. designed the study. B.H.,F.Z. W.P., Z.Z. and X.L.performed the experiments. B.H., F.Z., Z.Z.and X.L. counted and analyzed the experiment results. B.H.,F.Z., Q.K. and Y.Z. wrote the paper.

\section*{Competing interests}
The authors declare no competing interests.

\end{document}